\documentclass{svproc}
\usepackage{url}
\usepackage{biblatex}
\usepackage[pdftex]{graphicx}
\usepackage{tabularx,booktabs}
\usepackage[toc,page]{appendix}
\usepackage{multirow}
\usepackage{multicol}
\usepackage{graphicx}
\usepackage[dvipsnames]{xcolor}
\usepackage{import}
\usepackage{amsfonts}
\usepackage[algoruled,linesnumbered]{algorithm2e}
\usepackage{subcaption}

\usepackage{amsmath}

\newcommand{\argmin}{\mathop{\mathrm{argmin}}}
\newcommand{\argmax}{\mathop{\mathrm{argmax}}}

\usepackage{tikz}
\usetikzlibrary{arrows,positioning} 
\tikzset{
    >=stealth',
    punkt/.style={
           rectangle,
           rounded corners,
           draw=black, very thick,
           text width=8.5em,
           minimum height=2em,
           text centered},
    pil/.style={
           ->,
           thick,
           shorten <=2pt,
           shorten >=2pt,}
}

\begin{document}
\mainmatter              
\title{Optimizing the Neural Architecture of Reinforcement Learning Agents}
\titlerunning{Optimizing the Neural Architecture of Reinforcement Learning Agents
}  
%
\author{N. Mazyavkina\thanks{equal contribution}, S. Moustafa\footnotemark[1], I. Trofimov, E. Burnaev}
\authorrunning{N. Mazyavkina, S. Moustafa, I. Trofimov, E. Burnaev} 

%
\tocauthor{Nina Mazyavkina, Samir Moustafa, Ilya Trofimov, Evgeny Burnaev}
\institute{Skolkovo Institute of Science and Technology, Moscow, Russia,\\
\email{[n.mazyavkina, samir.mohamed, ilya.trofimov, e.burnaev]@skoltech.ru}}

\maketitle              

\begin{abstract}
Reinforcement learning (RL) enjoyed significant progress over the last years. One of the most important steps forward was the wide application of neural networks. However, architectures of these neural networks are quite simple and typically are constructed manually. In this work, we study recently proposed \textit{neural architecture search (NAS)} methods for optimizing the architecture of RL agents. We create two search spaces for the neural architectures and test two NAS methods: Efficient Neural Architecture Search (ENAS) and Single-Path One-Shot (SPOS). Next, we carry out experiments on the Atari benchmark and conclude that modern NAS methods find architectures of RL agents outperforming a manually selected one. 



\end{abstract}

\section{Introduction}
Over the last several years, deep learning (DL) has experienced enormous growth in popularity among the researchers from both the academia and the industry. 
Moreover, each of the separate tasks solved by the DL methods requires its own approach, one of the most important aspects of which is the choice of the neural network's (NN) architecture. In this case, it is essential to demonstrate good expertise and experience in the problem's field.
However, even then, the chosen architecture may not give any acceptable results until various heuristics and tricks will be applied to its construction. This motivated the emergence of the \textit{neural architecture search} (NAS) field, which focuses on automating the ways to find the optimal architecture for the specific tasks. 

Another family of methods that has been gaining popularity is reinforcement learning (RL) and deep reinforcement learning (deep RL), in particular. It consolidates a vast collection of machine learning methods, designed to solve a variety of Markov-Decision-Process-like problems. 
Over the last several years the successes of RL and deep RL has been frequently demonstrated by the research community: from better-than-human performance in ATARI \cite{mnih2013playing}, DOTA 2 \cite{mccandlish2018empirical}, Go \cite{silver2017mastering} to robotic manipulation \cite{gu2017deep}. 
In deep RL, neural networks are usually used to approximate a \textit{value} function, in the case of the value-based methods, or a \textit{policy} function, in the case of policy gradient methods. Moreover, actor-critic RL algorithms \cite{sutton2000policy,mnih2016asynchronous} combine these two NN approximations, in order to gain even better performance. Consequently, finding a suitable NN architecture is also a vital part of designing RL experiments. 

In this work, we are going to explore deep RL as a new application of NAS, i.e. deriving well-performing NN architectures for RL tasks. 
The motivation for the research in this direction is the following:
\begin{enumerate}
\item Only a single NN architecture is often chosen for many common benchmarks such as ATARI \cite{mnih2013playing} and MuJoCo \cite{todorov2012mujoco}, despite of them consisting of a big number of different environments. Hence, automatically finding a suitable network for each of the environments may lead to better results;

\item NAS can be useful in the cases of more complicated environments with bigger state and action spaces, where a more complicated deeper network might be required.
\end{enumerate}

Early NAS methods \cite{zoph2016neural, zoph2018learning} required training of numerous neural architectures. However, even training of one RL agent takes a significant amount of time. 
In this paper, we limit ourselves to fast \textit{one-shot} methods, which perform architecture search in the time not significantly larger than the training time of a single neural network. 
Most of the NAS methods were developed for computer vision applications, the major part -- for the object classification problem. 
At the same time, reinforcement learning is quite a different problem. The performance of an RL agent, that is, average reward, is not differentiable like the cross-entropy of object classification. 
Thus, only few popular NAS methods are suited for RL. In this work, we evaluate ENAS \cite{zoph2018learning} and SPOS \cite{guo2019single}.

The contribution of our paper is the following: we experimentally prove that modern one-shot NAS methods can be successfully applied for optimizing the neural architecture of RL agents. 
The source code is publicly available from \textit{\url{https://github.com/NinaMaz/NAS_RL_torch}}.

\section{Related work}
Early neural architecture search (NAS) approaches
treated this problem as a black-box optimization, that is, search over a discrete domain of architectures. Such methods are quite general but require training of numerous architectures and vast computational resources. One of the first proposed methods of this kind \cite{zoph2016neural, zoph2018learning} used reinforcement learning for the optimization process itself. Architecture creation, layer by layer, was done by an RL agent. Thus, the reward was the performance of the constructed network.
Other works proposed evolutionary optimization \cite{real2019regularized}, bayesian optimization based on Gaussian processes \cite{kandasamy2018neural}, bayesian performance predictors based on architecture features \cite{white2019bananas, shi2019multi}. Black-box optimization enjoy speedup from multi-fidelity methods \cite{trofimov2020multi}. Several benchmarks for NAS were developed \cite{ying2019bench, klyuchnikov2020bench, dong2020bench}.

The later family of methods -- \textit{one-shot NAS} -- gone beyond black-box optimization and utilized the structure a neural network. These methods involve the \textit{supernetwork}, which contain all the architectures from the search space as its subnetworks. Thus, all the architectures share weights of some of the blocks.
The architecture search in the supernetwork is performed simultaneously with the training of networks themselves. 
The one-shot methods are: ENAS \cite{pham2018efficient}, numerous modifications of DARTS \cite{liu2018darts, xu2019pc, chen2019progressive, liang2019darts+, dong2019searching, cai2018proxylessnas}, single path one-shot \cite{guo2019single}, random search with weight sharing \cite{li2019random, bender2019understanding}.

Most of the existing research focuses on problems from computer vision and linguistics.
There are no papers about applications of modern NAS methods to RL to the best of our knowledge.


\section{Reinforcement Learning Methods}

In our experiments, we have used reinforcement learning for training both ENAS controller and sampled child networks. On the other hand, SPOS does not use a trainable controller for architecture sampling and, hence, the RL methods, mentioned in this section, do not concern it. 

An LSTM controller, used in the ENAS framework, is trained with  REINFORCE \cite{Williams:92} algorithm. REINFORCE belongs to a group of policy-based methods, which focuses on the straightforward approximation of the optimal policy, via calculating the direct gradient of the parameterized objective function $J$:
 $$\nabla_\theta J_\theta = \mathbb{E}_t\Big(\nabla_\theta \log \pi_\theta (a_t | s_t) R_t\Big).$$
 In our case, $\theta$ are the parameters of the neural network, outputting the logits, from which the actions $a_t$ can be derived; $\pi$ is the policy, $s_t$ are the states, $R_t$ is the sum of the discounted rewards collected so far. 
 Specifically, in the case of REINFORCE algorithm, the update to the parameters $\theta$ takes the form:
 $$
 \theta \leftarrow \theta + \gamma^{t}R_t\nabla_\theta \log \pi_\theta (a_t | s_t),
 $$
 where $\gamma$ is the discount factor.
  In order to reduce the variance of the gradient estimation from the formula above, we subtract the moving average baseline from the discounted reward function.
 
In terms of the training process of the child networks, we use another policy-based method - Proximal Policy Optimization (PPO) \cite{schulman2017proximal} to update their parameters. The objective function for PPO has the following form:

$$J_\theta^{PPO} = \mathbb{E}_t\Big(\min\big[{ratio_t(\theta)A_t, clip(ratio_t(\theta), 1+\epsilon, 1-\epsilon)A_t\big]\Big)},$$
where $A_t$ is the advantage function, $ratio$ is the probability ratio under the new and old policies, $\epsilon$ is the clipping parameter.

\section{Neural Architecture Search Methods}

\begin{figure}
 \centering 
 \includegraphics[width=0.5\textwidth]{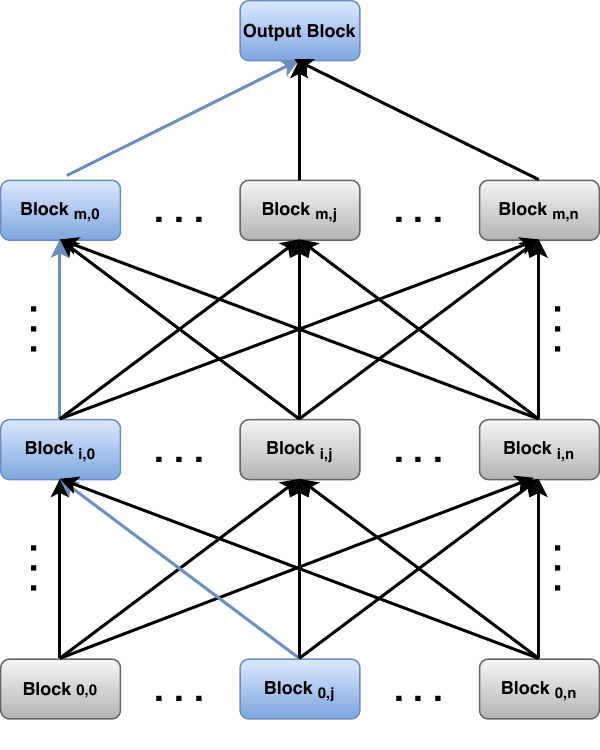}
  \caption{Supernetwork architecture for a generic design. Each block can be a part of neural network, and each connection between sequence of blocks can be a complete architecture. The \textcolor{blue}{blue} path represent a complete architecture.
  }
  \label{fig:supernet_single}
\end{figure}

\subsection{Adaptation to Reinforcement Learning}
\label{sec:adapation}
Most of the existing research on NAS is dedicated to computer vision (particularly, object classification) or computational linguistics applications. We adapt the existing one-shot NAS methods to reinforcement learning. One-shot methods assume the \textit{supernetwork} which contain all the architectures from the search space as its subgraphs (fig. \ref{fig:supernet_single}). All the architectures share weights of some of the blocks. 
That is, each layer in the supernetwork $i$ has a list of $Block_{i}$ options, only one option can be selected for a particular subnetwork. Such simplification was proposed to reduce the co-adaptation between blocks. The whole search space is $\mathcal{A} = Block_0 \times \ldots \times Block_n$. Some initial or final layers of the supernetwork can be fixed and not contain choice blocks.

One-shot methods typically contain a \textit{fitting} stage. During the \textit{fitting} stage, the subnetwork $\alpha \in \mathcal{A}$ is sampled by some rule and its weights $\Theta(\alpha)$ are updated by a SGD-like step for a batch of data $B$
\begin{equation}
\label{eq:cv-sgd}
\Theta(\alpha) \gets \Theta(\alpha) - \eta \nabla_{\Theta(\alpha)} \sum_{i\in B} \ell(y_i, N(\alpha, \Theta(\alpha), x_i)),
\end{equation}
where $\ell(\cdot)$ is the loss function, $N(\alpha, \theta, x_i)$ is the network of the architecture $\alpha$ having weights $\Theta(\alpha)$.
The \textit{evaluation} of the architecture $\alpha$ typically involves calculation of performance (accuracy) on the validation dataset $D_{val}$
\begin{equation}
\label{eq:cv-acc}
\frac{1}{|D_{val}|}\sum_{i \in D_{val}}[y_i = N(\alpha, \Theta(\alpha),x_i)].
\end{equation}
We adapt one-shot methods to RL in the following way. In our experiments, the neural network $N(\alpha, \Theta(\alpha), x_i)$ corresponds to a policy $\pi_{\Theta(\alpha)}$. Instead of SGD-like step (\ref{eq:cv-sgd}) we do the step of PPO 
\begin{equation}
\Theta(\alpha) \gets \Theta(\alpha) - \eta \nabla_{\Theta(\alpha)} J^{PPO}_{\Theta(\alpha)}.
\end{equation}
The performance of the network is estimated by 
$\mathbb{E}_{t} [ R_{t}[\pi_{\Theta(\alpha)}]$
instead of (\ref{eq:cv-acc}).

\subsection{Efficient Neural Architecture Search (ENAS)}
\begin{figure}
\centering
\begin{tikzpicture}[node distance=1cm, auto,]
 \node[] (dummy) {};
 \node[punkt,right=of dummy] (t) {Child network};
 \node[punkt,left=of dummy] (g) {Controller}
   edge[pil,->, bend left=45] node[auto] {Action - sampling a child architecture} (t)
   edge[pil,<-, bend right=45] node[auto, below] {Reward - the accuracy of the trained child network} (t);
\end{tikzpicture}

\caption{A general framework of Neural Architecture Search (NAS)}
\label{controller_pic}
\end{figure}
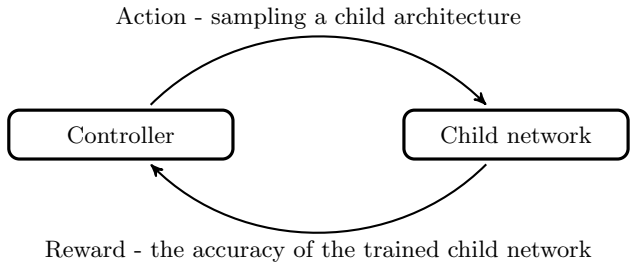
Efficient Neural Architecture Search (ENAS)
\cite{pham2018efficient} 
is a NAS method, used in our experiments to find a well-performing neural network in the ATARI games environment. ENAS consists of a controller, which samples child models, a search strategy and an evaluation strategy.  

In the case of ENAS, the controller is an LSTM network that outputs a one-hot-encoded architecture of a child model. The controller's inputs are the previous step's architecture and the reward it has received. The possible choices for a set of child architectures are determined by the search space. The particular variants of the search spaces that we have used are covered in section \ref{sec:search_sp}.

The authors of \cite{pham2018efficient,zoph2016neural} have proposed to train an ENAS controller by using RL. In our experiments, we have employed a REINFORCE algorithm, which has also been used in the original paper to update the controller's network parameters.


In the original paper, due to the different nature of the problems ENAS has been tested on, the child networks, sampled by the controller, are trained until convergence. However, this becomes difficult when the RL environments are considered -- the agents usually require longer training times, and the stochasticity of the environments makes the training process much more unpredictable. In our work, we will demonstrate that despite the aforementioned problems, the number of child training timesteps, chosen for our experiments, is still sufficient to determine which networks are better-performing than others. 

Finally, the overall sequence of ``sampling a child architecture - training a child model - feeding the resulting reward to LSTM controller" (see fig. \ref{controller_pic}) defines a single epoch of ENAS training. In the previous works on iterative RL NAS methods \cite{zoph2016neural,zoph2018learning}, the fact that such an epoch will take a long time to compute, has limited these methods' practicality in regard to the real-life problems. The authors of ENAS, however, have come up with the solution to this problem by sharing the weights of all of the child models. This way, the ENAS becomes much more efficient than its predecessors, and, therefore, much more suitable for RL problems.

\subsection{Single-Path One-Shot with Uniform Sampling (SPOS)}

The method ``Single-path one-shot with uniform sampling'' was proposed in \cite{guo2019single}.
The SPOS method assumes two steps: 1) supernetwork fitting 2) best architecture selection.
The distinctive feature of SPOS is that subnetworks are sampled from the supernetwork uniformly at random.

Thus, weights $\Theta^*$ of the supernetwork are the solution of the following problem 
\[
\Theta^* = \argmin_{\Theta} \mathbb{E}_{\alpha\sim P} [ L_{train}(N(\alpha, \Theta(\alpha)))], 
\]
where $L_{train}(\cdot)$ is the train loss, $P$ - the uniform distribution.
During the architecture selection phase, the best subnetwork $\alpha^*$ is selected by the validation accuracy $Acc_{val}$
\begin{equation}
\label{eq:opt-val}
\alpha^* = \argmax_{\alpha \in \mathcal{A}} Acc_{val}(N(\alpha, \Theta^*(\alpha))).
\end{equation}
This step requires only inference for the validation data. 
In the original paper \cite{guo2019single}, an evolutionary optimization was used to solve (\ref{eq:opt-val}) since the search space was huge. Instead of evolutionary optimization, we do the full search since our search spaces are small.

The adaptation of SPOS to RL is done as described in the Section \ref{sec:adapation}. The subnetwork $N(\alpha, \Theta(\alpha))$ corresponds to a policy $\pi_{\Theta(\alpha)}$. 
Thus, SPOS for optimizing the neural architecture of the RL agent solves the following problem
\begin{align}
\Theta^* = \argmax_{\Theta} \mathbb{E}_{\alpha \sim P} \mathbb{E}_{t} [ R_{t} [\pi_{\Theta(\alpha)}]], \\
\alpha^* = \argmax_{\alpha \in \mathcal{A}} \mathbb{E}_{t} [ R_{t}[\pi_{\Theta^*(\alpha)}].
\end{align}

\section{Experiments}

\subsection{Atari Environment}
\label{sec:atari}

In our experiments, we used the Open AI Gym framework \cite{Brockman2016OpenAIG}, particularly --  \textit{Breakout} and \textit{Freeway} Atari environments.
We chose the \textit{Breakout} because it’s a popular benchmark, having moderate standard deviation of RL agent's reward ($401.2 \pm 26.9$, \cite{Humanlevel2015}). 
In opposite to the \textit{Breakout}, the \textit{Freeway} environment has very low relative standard deviation of reward ($30.3 \pm 0.7$, \cite{Humanlevel2015}).
The reward of RL agent with random behavior for \textit{Breakout} and \textit{Freeway} is nearly zero so we can make sure that our policy network makes non-stochastic behavior.

We have trained the child networks in the manner described in \cite{espeholt2018impala}, i.e., we have used 8 agents, sharing a policy, trained simultaneously in 8 environments with PPO, in order to collect the trajectories for the policy update. Each of these agents is trained for 128 steps. After that, the controller collects the architecture's rewards. The controller's policy is updated every ten steps using the REINFORCE algorithm. Overall, the number of training steps for one experiment equals 10 million. More implementation details are in Appendix \ref{app:hyperparams}.

It is important to note that the 'scratch' experimental results demonstrate lower reward values than the ones reported in the original PPO paper \cite{schulman2017proximal}. This is due to the fact that in our experiments we have used a smaller number of training timesteps than the classical version of PPO (10M vs. 40M). The reason for this has been that the main aim of our research focuses on investigating whether NAS has a positive effect on RL training process overall, and not on beating the existing ATARI baselines.
\subsection{Search Spaces}
\label{sec:search_sp}

\begin{figure}
\begin{subfigure}{\textwidth}
 \centering 
 \includegraphics[width=\textwidth]{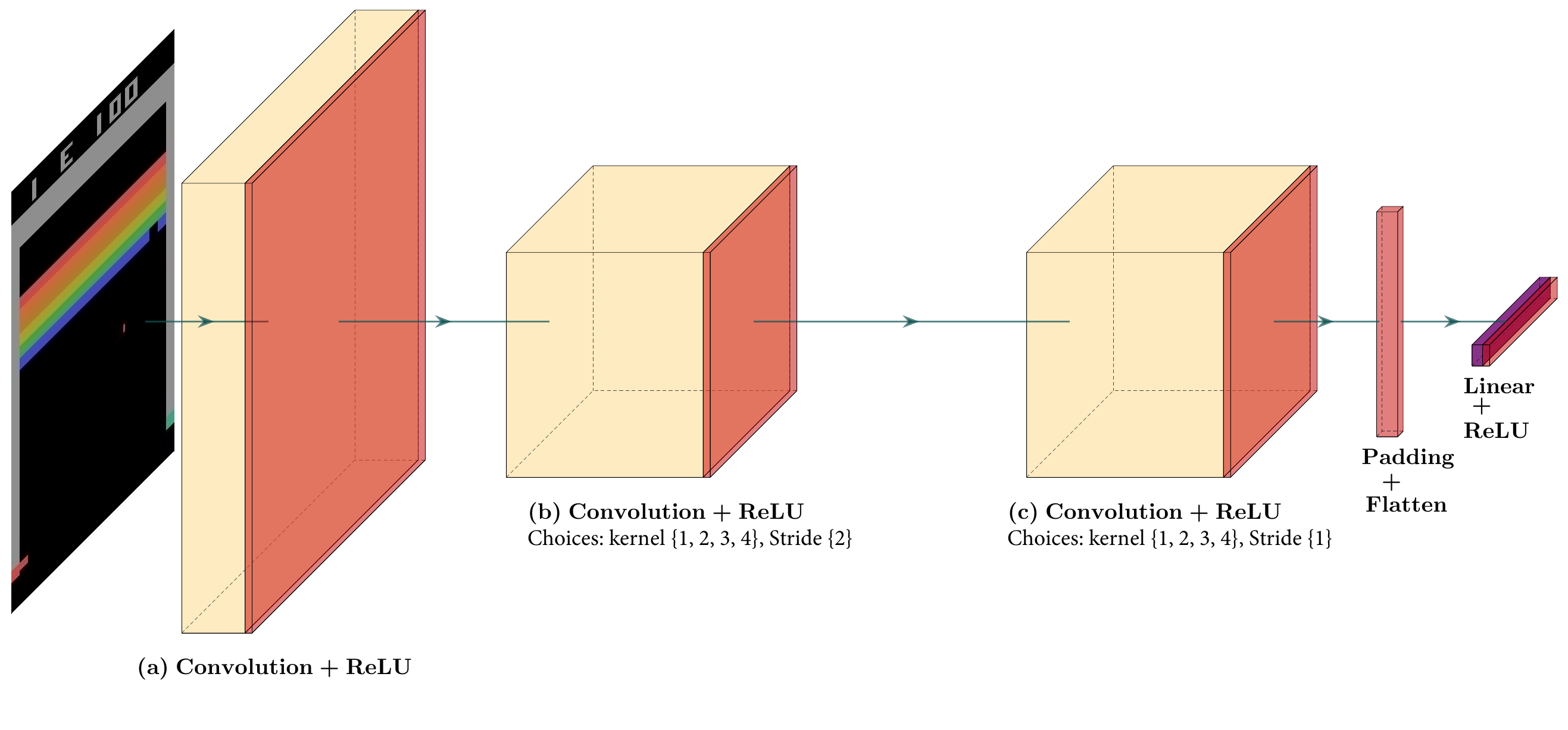}
  \caption{\textbf{Search space 1:} Block(a): fixed kernel size:{8}, fixed stride:{4};  Block(b) and Block(c): kernel size is chosen between {1, 2, 3, 4, and 5}, and fixed stride: {2, 1} respectively.}
  \label{fig:architecture_1}
\end{subfigure}
\begin{subfigure}{\textwidth}
 \centering 
 \includegraphics[width=\textwidth]{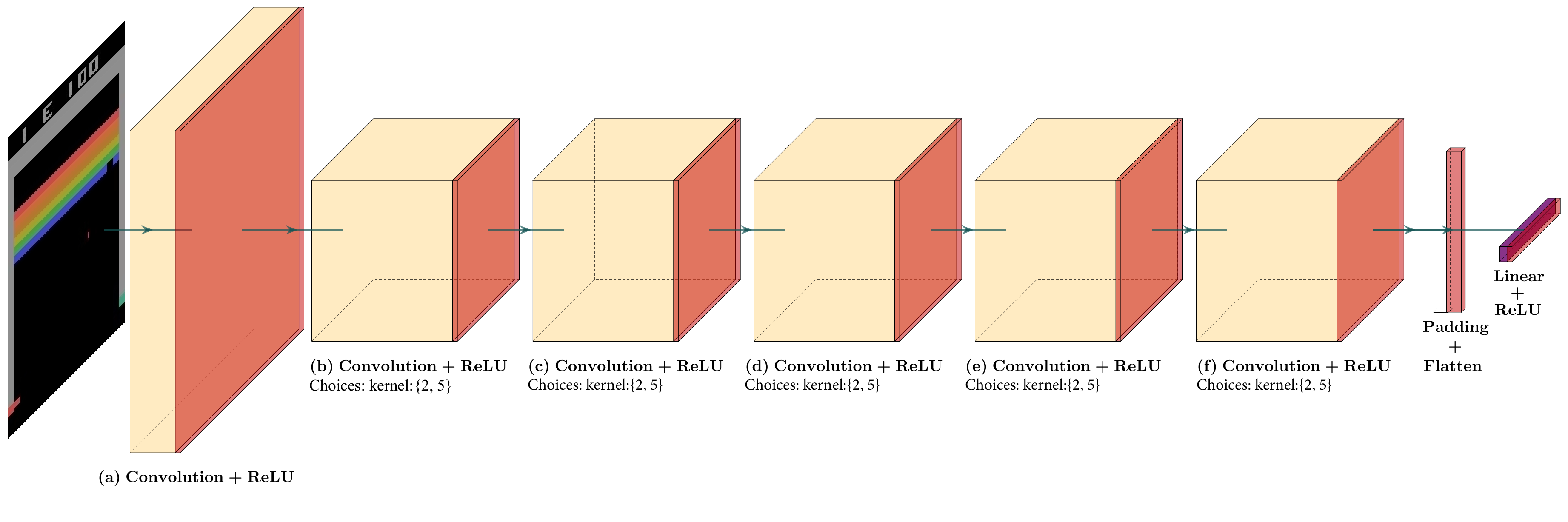}
  \caption{\textbf{Search space 2:} Block(a): fixed kernel size:{8}, fixed stride:{4};  Block(b)- Block(f): kernel size is chosen between {2 and 5}, with fixed stride: {1} for all of them.}
  \label{fig:architecture_2}
\end{subfigure}
\caption{Search spaces used in experiments.}
\end{figure}

The search spaces that we designed are the extension of the Nature-CNN architecture \cite{Humanlevel2015}, where convolutional layers are followed by
a linear layer which outputs the number of values equal to the size of our action space \cite{Humanlevel2015}. 


In order to facilitate the varying sizes of the layers' parameters and, hence, enable the weight sharing, we use the following techniques in the overall design of the network:
\begin{enumerate}
  \item \textbf{Convolution and max-pooling to the same size}:
  Map the input to one size every time after each convolutional and max-pooling layer. 
  \item \textbf{Padding to exact size}:
  Map the output of the last convolutional layer to the target size to be able to fix it during the flattening of the convolutions.
  
\end{enumerate}

In our experiments we have covered two architectural search spaces:

\textbf{Search space 1: 25 architectures}.
The architecture starts with a fixed convolutional layer with input channels equal to 4, and the output channels - to 32, kernel size - 8 and moving stride - 4.
This layer is followed by two convolutional layers 
with output channels equal to 64 per each and with strides 1 and 2. The choices for kernel sizes are 1, 2, 3, 4, 5. The size of this search space is $5^2$.

\textbf{Search space 2: 32 architectures}.
Instead of two convolutional layers (we do not take into account the first fixed convolutional layer, as we do not vary its kernel size) used in the search space 1, we have used 5 convolutional layers 
with output channels equal to 64, moving strides equal to 1. The choices for kernels are 2 and 5, which makes the size of this search space equal to $2^5$.

The experimental architecture spaces that we have used can be seen in Figure \ref{fig:architecture_1} and Figure \ref{fig:architecture_2}, and also in Appendix \ref{app:search-spaces}.
The search spaces do not contain very deep architectures, having depth 7 at most.

\subsection{Methodology}
Firstly, we trained all the architectures from scratch (implementation details are in Appendix \ref{app:hyperparams}). For all of those architectures we saved \textit{mean reward} and \textit{total reward} averaged over last 100 episodes. 
These metrics were used as a tabular benchmark, eliminating the need to train the same architecture multiple times.
We used these metrics later to compare the performances of the found architectures by various NAS methods.

Both NAS methods under evaluation (ENAS, SPOS) share the same high-level structure: 
\begin{enumerate}
\item Run neural architecture search method for the given search space $\mathcal{A}$;
\item Select top-K architectures from the search space $\mathcal{A}$ by a \textit{proxy performance};
\item Train these top-K architectures from scratch;
\item Return the best one by the \textit{true performance}. \end{enumerate}

In our experiments, we selected top-3 architectures for the methods under evaluation. We repeated the search 4 times with different seeds and averaged results.

The \textit{proxy performance} is a part of the NAS method, and it is fast to calculate. The calculation time of the proxy performance is negligible to the time of RL agent training from scratch. For ENAS, the proxy performance of an architecture is the probability of sampling this architecture by the controller. For SPOS, the proxy performance of an architecture is calculated from weights of this architecture in the supernetwork. Namely, the proxy performance is the mean reward of an agent with corresponding weights. 

The \textit{true performance} is the total/mean reward of an RL agent trained from scratch.

\subsection{Random search baseline} 
As a simple baseline, we used the following random search algorithm:
\begin{enumerate}
\item Select K architectures from the search space $\mathcal{A}$ at random;
\item Train these K architectures from scratch;
\item Return the best one by the \textit{true performance}.
\end{enumerate}

As for ENAS and SPOS, we selected top-3 architectures.
The only difference with ENAS/SPOS methods is that architectures are selected at random instead of by \textit{proxy performance}.
We estimated the variance of the random search by repeating it for 1000 times.




\subsection{Experiments with larger search spaces}
We have also tried to expand the search space by increasing the number of consecutive convolutional layers up to 9 and choosing a suitable kernel size and a number of output channels for each of them. However, the results were close to the ones received by random search, which can be caused by the fact that NAS can experience problems with increasingly complex search spaces. For that reason, we do not present the results of these experiments in the paper. 


\begin{table}[t]
\centering 
 \caption{Reward mean and total reward of RL agents with various architectures. Each score is the mean of last 100 episodes.}
 \label{tbl:results_table}
\begin{tabularx}{\textwidth}{@{}lccccc@{}}
\toprule

            & & \multicolumn{2}{c}{Search space 1}  &  \multicolumn{2}{c}{Search space 2} \\ 
\midrule

            & & Breakout & Freeway & Breakout & Freeway \\ 
\midrule

\multirow{2}{*}{Random Search}      & reward mean     & 54.7 $\pm$ 8.3    & 28.4 $\pm$ 1.0  & 33.1 $\pm$ 29.5   & 21.6 $\pm$ 0.2  \\ 
                                    & total reward    & 147.2 $\pm$ 25.5  & \textbf{31.7 $\pm$ 0.8}  & 105.7 $\pm$ 94.2    & \textbf{22.0 $\pm$ 0.2}  \\ 
\addlinespace

 \multirow{1}{*}{Nature-CNN \cite{Humanlevel2015},} & reward mean               & 57.1          & 13.1      & -         & -     \\ 
 \quad reproduced                                   & total reward              & 157.9         & 19.0      & -         & -     \\ 

\addlinespace

\multirow{2}{*}{ENAS}   & reward mean             & \textbf{61.4 $\pm$ 1.8}          & 26.4 $\pm$ 1.1  & 30.7 $\pm$ 21.7      & 21.5 $\pm$ 0.1                     \\
                        & total reward             & \textbf{161.1 $\pm$ 9.8}    & 30.7 $\pm$ 1.3    & 91.4  $\pm$ 64.4   & \textbf{22.0 $\pm$ 0.2}                     \\ 
\addlinespace

\multirow{2}{*}{SPOS}   & reward mean               & 39.7 $\pm$ 18.6         & \textbf{29.6 $\pm$ 0.8}    & \textbf{39.9 $\pm$ 41.0}     & \textbf{21.7 $\pm$ 0.1}     \\ 
                        & total reward              & 144.4 $\pm$ 55.0         & 29.4 $\pm$ 5.0             & \textbf{180.6 $\pm$ 72.5}    & \textbf{22.0 $\pm$ 0.1}              \\ 

\bottomrule
\end{tabularx}
\end{table}

\begin{figure}[t]
\centering
\begin{subfigure}{0.45\linewidth}
    \includegraphics[width=\linewidth]{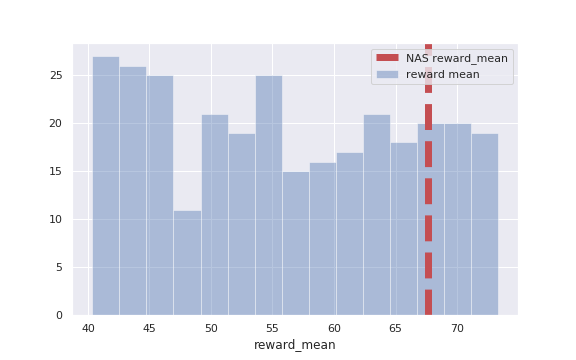}
    \caption{Search Space 1: Breakout}
\end{subfigure}
\hfil
\begin{subfigure}{0.45\linewidth}
    \includegraphics[width=\linewidth]{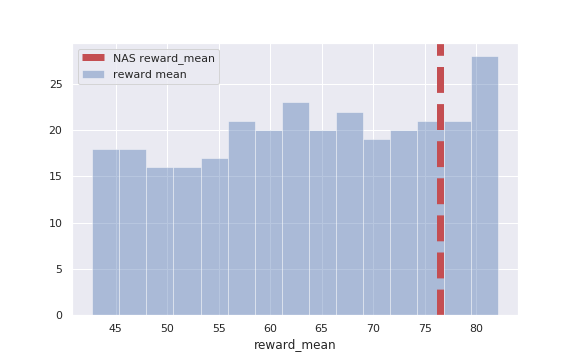}
    \caption{Search Space 2: Breakout}
\end{subfigure}

\begin{subfigure}{0.45\linewidth}
    \includegraphics[width=\linewidth]{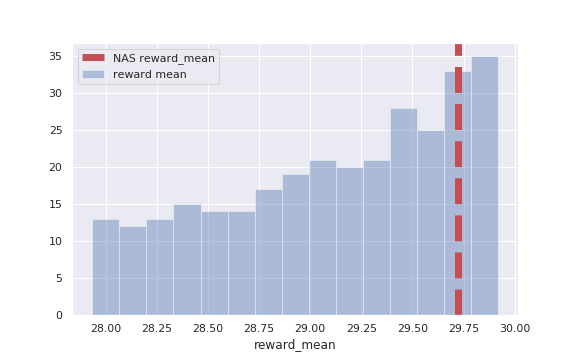}
    \caption{Search Space 1: Freeway}
\end{subfigure}
\hfil
\begin{subfigure}{0.45\linewidth}
    \includegraphics[width=\linewidth]{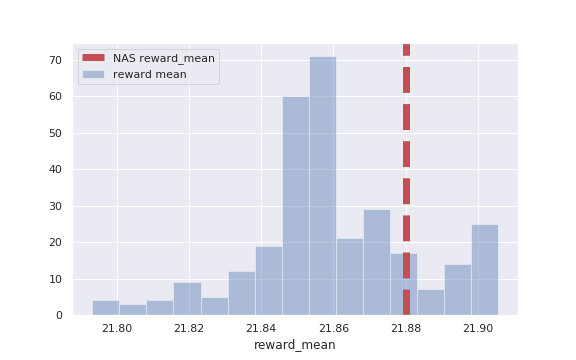}
    \caption{Search Space 2: Freeway}
\end{subfigure}
\caption{Histograms for the reward mean of RL agents having different architectures from search spaces. The red vertical line depicts the best architecture found by NAS methods.}
\label{fig:histograms}
\end{figure}

\section{Discussion}
Table \ref{tbl:results_table} shows the results of our experiments.
For the Breakout environment, ENAS performs better than random search on the search space 1, while SPOS performs better on the search space 2. For the Freeway environment, there is no clear benefit of NAS methods. Variance of the both of the methods are quite high.
At the same time, SPOS is simpler since it does not contain an auxiliary controller network. 

It is interesting to compare the performances of found architectures with the performance of manually selected Nature-CNN architecture \cite{Humanlevel2015} (see description in the Appendix \ref{app:nature-CNN-architecture}). The Nature-CNN architecture belongs to the search space 1. For the fair comparison, in Table \ref{tbl:results_table}, we report the rewards after training of the Nature-CNN architecture with our pipeline (Section \ref{sec:atari}). The architectures found by NAS methods outperform the manually selected Nature-CNN by a considerable margin.

The existing research on RL methods typically uses the same architectures for different environments.
However, we found that this is not optimal. Different architectures are optimal for different environments, see Appendix \ref{app:found-architectures}.

Figure \ref{fig:histograms} shows histograms of RL agents' mean rewards from different search spaces. The red vertical lines depict the best architecture found by NAS methods. We conclude that NAS methods can find top architectures in both of the search spaces. 


\section{Conclusion}
Traditionally, the progress in RL field came mostly from the development of new methods. Neural architectures of RL agents remained relatively simple when compared to computer vision applications. 

In this paper, we have applied modern neural architecture search methods for optimizing the architecture of RL agents. We have evaluated ENAS \cite{zoph2018learning} and SPOS \cite{guo2019single} methods. Both of them found better architectures than manually picked by experts. We suppose that many RL application can benefit from using better neural architectures.
Testing NAS methods for larger search spaces is an interesting topic for further research.

\subsection*{Acknowledgments}
Authors are thankful to Mikhail Konobeeb.


\nocite{*}
\printbibliography

@Article{Humanlevel2015,
  Title                     = {Human-level control through deep reinforcement learning},
  Author                    = {Mnih, Volodymyr and Kavukcuoglu, Koray and Silver, David and  Rusu, A and Veness,Joel },
  Year                      = {2015},
  Doi                       = {10.1038/nature14236},
  Pages                     = {2--3}
}

@Article{Girosi1995,
  Title                     = {Regularization Theory and Neural Networks Architectures},
  Author                    = {Girosi, Federico and Jones,Michael and Poggio, Tomaso},
  Year                      = {1995},
  Volume                    = {7},
  Doi                       = {10.1162/neco.1995.7.2.219},
  ISSN                      = {0899-7667}
}

@Article{Schrittwieser2019MasteringAG,
  Title                     = {Mastering Atari, Go, Chess and Shogi by Planning with a Learned Model},
  Author                    = {Julian Schrittwieser and Ioannis Antonoglou and Thomas Hubert and Karen Simonyan and Laurent Sifre and Simon Schmitt and Arthur Guez and Edward Lockhart and Demis Hassabis and Thore Graepel and Timothy P. Lillicrap and David Silver},
  Journal                   = {ArXiv},
  Year                      = {2019},
  Volume                    = {abs/1911.08265}
}

@Article{kapturowski2018recurrent,
    Title                   = {Recurrent Experience Replay in Distributed Reinforcement Learning},
    Author                  = {Steven Kapturowski and Georg Ostrovski and Will Dabney and John Quan and Remi Munos},
    booktitle               = {International Conference on Learning Representations},
    Year                    = {2019},
    url                      = {https://openreview.net/forum?id=r1lyTjAqYX},
    }

@Article{Brockman2016OpenAIG,
  Title                 = {OpenAI Gym},
  Author                = {Greg Brockman and Vicki Cheung and Ludwig Pettersson and Jonas Schneider and John Schulman and Jie Tang and Wojciech Zaremba},
  journal               ={ArXiv},
  year                  ={2016},
  volume                ={abs/1606.01540}
}

@article{mnih2013playing,
  title={Playing atari with deep reinforcement learning},
  author={Mnih, Volodymyr and Kavukcuoglu, Koray and Silver, David and Graves, Alex and Antonoglou, Ioannis and Wierstra, Daan and Riedmiller, Martin},
  journal={arXiv preprint arXiv:1312.5602},
  year={2013}
}

@article{mccandlish2018empirical,
  title={An empirical model of large-batch training},
  author={McCandlish, Sam and Kaplan, Jared and Amodei, Dario and Team, OpenAI Dota},
  journal={arXiv preprint arXiv:1812.06162},
  year={2018}
}

@article{silver2017mastering,
  title={Mastering the game of go without human knowledge},
  author={Silver, David and Schrittwieser, Julian and Simonyan, Karen and Antonoglou, Ioannis and Huang, Aja and Guez, Arthur and Hubert, Thomas and Baker, Lucas and Lai, Matthew and Bolton, Adrian and others},
  journal={Nature},
  volume={550},
  number={7676},
  pages={354--359},
  year={2017},
  publisher={Nature Publishing Group}
}

@inproceedings{gu2017deep,
  title={Deep reinforcement learning for robotic manipulation with asynchronous off-policy updates},
  author={Gu, Shixiang and Holly, Ethan and Lillicrap, Timothy and Levine, Sergey},
  booktitle={2017 IEEE international conference on robotics and automation (ICRA)},
  pages={3389--3396},
  year={2017},
  organization={IEEE}
}

@inproceedings{sutton2000policy,
  title={Policy gradient methods for reinforcement learning with function approximation},
  author={Sutton, Richard S and McAllester, David A and Singh, Satinder P and Mansour, Yishay},
  booktitle={Advances in neural information processing systems},
  pages={1057--1063},
  year={2000}
}

@inproceedings{mnih2016asynchronous,
  title={Asynchronous methods for deep reinforcement learning},
  author={Mnih, Volodymyr and Badia, Adria Puigdomenech and Mirza, Mehdi and Graves, Alex and Lillicrap, Timothy and Harley, Tim and Silver, David and Kavukcuoglu, Koray},
  booktitle={International conference on machine learning},
  pages={1928--1937},
  year={2016}
}

@inproceedings{todorov2012mujoco,
  title={Mujoco: A physics engine for model-based control},
  author={Todorov, Emanuel and Erez, Tom and Tassa, Yuval},
  booktitle={2012 IEEE/RSJ International Conference on Intelligent Robots and Systems},
  pages={5026--5033},
  year={2012},
  organization={IEEE}
}

@article{bellman1952theory,
  title={On the theory of dynamic programming},
  author={Bellman, Richard},
  journal={Proceedings of the National Academy of Sciences of the United States of America},
  volume={38},
  number={8},
  pages={716},
  year={1952},
  publisher={National Academy of Sciences}
}

@book{sutton2018reinforcement,
  title={Reinforcement learning: An introduction},
  author={Sutton, Richard S and Barto, Andrew G},
  year={2018},
  publisher={MIT press}
}

@article{levine2020offline,
  title={Offline reinforcement learning: Tutorial, review, and perspectives on open problems},
  author={Levine, Sergey and Kumar, Aviral and Tucker, George and Fu, Justin},
  journal={arXiv preprint arXiv:2005.01643},
  year={2020}
}

@misc{schulman2017proximal,
    title={Proximal Policy Optimization Algorithms},
    author={John Schulman and Filip Wolski and Prafulla Dhariwal and Alec Radford and Oleg Klimov},
    year={2017},
    eprint={1707.06347},
    archivePrefix={arXiv},
    primaryClass={cs.LG}
}

@article{Williams:92,
  author = {Williams, R. J.},
  journal = {Machine Learning},
  title = {Simple statistical gradient-following algorithms for connectionist reinforcement learning},
  year = 1992
}

@article{pham2018efficient,
  title={Efficient neural architecture search via parameter sharing},
  author={Pham, Hieu and Guan, Melody Y and Zoph, Barret and Le, Quoc V and Dean, Jeff},
  journal={arXiv preprint arXiv:1802.03268},
  year={2018}
}

@article{zoph2016neural,
  title={Neural architecture search with reinforcement learning},
  author={Zoph, Barret and Le, Quoc V},
  journal={arXiv preprint arXiv:1611.01578},
  year={2016}
}

@inproceedings{zoph2018learning,
  title={Learning transferable architectures for scalable image recognition},
  author={Zoph, Barret and Vasudevan, Vijay and Shlens, Jonathon and Le, Quoc V},
  booktitle={Proceedings of the IEEE conference on computer vision and pattern recognition},
  pages={8697--8710},
  year={2018}
}

@article{liu2018darts,
  title={Darts: Differentiable architecture search},
  author={Liu, Hanxiao and Simonyan, Karen and Yang, Yiming},
  journal={arXiv preprint arXiv:1806.09055},
  year={2018}
}

@inproceedings{real2019regularized,
  title={Regularized evolution for image classifier architecture search},
  author={Real, Esteban and Aggarwal, Alok and Huang, Yanping and Le, Quoc V},
  booktitle={Proceedings of the aaai conference on artificial intelligence},
  volume={33},
  pages={4780--4789},
  year={2019}
}

@article{adam2019understanding,
  title={Understanding neural architecture search techniques},
  author={Adam, George and Lorraine, Jonathan},
  journal={arXiv preprint arXiv:1904.00438},
  year={2019}
}

@article{elsken2018efficient,
  title={Efficient multi-objective neural architecture search via lamarckian evolution},
  author={Elsken, Thomas and Metzen, Jan Hendrik and Hutter, Frank},
  journal={arXiv preprint arXiv:1804.09081},
  year={2018}
}

@article{cai2018proxylessnas,
  title={Proxylessnas: Direct neural architecture search on target task and hardware},
  author={Cai, Han and Zhu, Ligeng and Han, Song},
  journal={arXiv preprint arXiv:1812.00332},
  year={2018}
}

@article{xu2019pc,
  title={Pc-darts: Partial channel connections for memory-efficient differentiable architecture search},
  author={Xu, Yuhui and Xie, Lingxi and Zhang, Xiaopeng and Chen, Xin and Qi, Guo-Jun and Tian, Qi and Xiong, Hongkai},
  journal={arXiv preprint arXiv:1907.05737},
  year={2019}
}

@inproceedings{chen2019progressive,
  title={Progressive differentiable architecture search: Bridging the depth gap between search and evaluation},
  author={Chen, Xin and Xie, Lingxi and Wu, Jun and Tian, Qi},
  booktitle={Proceedings of the IEEE International Conference on Computer Vision},
  pages={1294--1303},
  year={2019}
}

@article{lindauer2019best,
  title={Best practices for scientific research on neural architecture search},
  author={Lindauer, Marius and Hutter, Frank},
  journal={arXiv preprint arXiv:1909.02453},
  year={2019}
}

@inproceedings{kandasamy2018neural,
  title={Neural architecture search with bayesian optimisation and optimal transport},
  author={Kandasamy, Kirthevasan and Neiswanger, Willie and Schneider, Jeff and Poczos, Barnabas and Xing, Eric P},
  booktitle={Advances in Neural Information Processing Systems},
  pages={2016--2025},
  year={2018}
}

@article{liang2019darts+,
  title={Darts+: Improved differentiable architecture search with early stopping},
  author={Liang, Hanwen and Zhang, Shifeng and Sun, Jiacheng and He, Xingqiu and Huang, Weiran and Zhuang, Kechen and Li, Zhenguo},
  journal={arXiv preprint arXiv:1909.06035},
  year={2019}
}

@article{yang2019evaluation,
  title={NAS evaluation is frustratingly hard},
  author={Yang, Antoine and Esperan{\c{c}}a, Pedro M and Carlucci, Fabio M},
  journal={arXiv preprint arXiv:1912.12522},
  year={2019}
}

@article{white2019bananas,
  title={BANANAS: Bayesian Optimization with Neural Architectures for Neural Architecture Search},
  author={White, Colin and Neiswanger, Willie and Savani, Yash},
  journal={arXiv preprint arXiv:1910.11858},
  year={2019}
}

@article{guo2019single,
  title={Single path one-shot neural architecture search with uniform sampling},
  author={Guo, Zichao and Zhang, Xiangyu and Mu, Haoyuan and Heng, Wen and Liu, Zechun and Wei, Yichen and Sun, Jian},
  journal={arXiv preprint arXiv:1904.00420},
  year={2019}
}

@inproceedings{dong2019searching,
  title={Searching for a robust neural architecture in four gpu hours},
  author={Dong, Xuanyi and Yang, Yi},
  booktitle={Proceedings of the IEEE Conference on Computer Vision and Pattern Recognition},
  pages={1761--1770},
  year={2019}
}

@article{shi2019multi,
  title={Multi-objective Neural Architecture Search via Predictive Network Performance Optimization},
  author={Shi, Han and Pi, Renjie and Xu, Hang and Li, Zhenguo and Kwok, James T and Zhang, Tong},
  journal={arXiv preprint arXiv:1911.09336},
  year={2019}
}

@article{li2019random,
  title={Random search and reproducibility for neural architecture search},
  author={Li, Liam and Talwalkar, Ameet},
  journal={arXiv preprint arXiv:1902.07638},
  year={2019}
}

@article{bender2019understanding,
  title={Understanding and simplifying one-shot architecture search},
  author={Bender, Gabriel},
  year={2019}
}

@article{trofimov2020multi,
  title={Multi-fidelity Neural Architecture Search with Knowledge Distillation},
  author={Trofimov, Ilya and Klyuchnikov, Nikita and Salnikov, Mikhail and Filippov, Alexander and Burnaev, Evgeny},
  journal={arXiv preprint arXiv:2006.08341},
  year={2020}
}

@article{klyuchnikov2020bench,
  title={NAS-Bench-NLP: Neural Architecture Search Benchmark for Natural Language Processing},
  author={Klyuchnikov, Nikita and Trofimov, Ilya and Artemova, Ekaterina and Salnikov, Mikhail and Fedorov, Maxim and Burnaev, Evgeny},
  journal={arXiv preprint arXiv:2006.07116},
  year={2020}
}

@inproceedings{ying2019bench,
  title={Nas-bench-101: Towards reproducible neural architecture search},
  author={Ying, Chris and Klein, Aaron and Christiansen, Eric and Real, Esteban and Murphy, Kevin and Hutter, Frank},
  booktitle={International Conference on Machine Learning},
  pages={7105--7114},
  year={2019}
}

@article{dong2020bench,
  title={Nas-bench-102: Extending the scope of reproducible neural architecture search},
  author={Dong, Xuanyi and Yang, Yi},
  journal={arXiv preprint arXiv:2001.00326},
  year={2020}
}

@article{espeholt2018impala,
  title={Impala: Scalable distributed deep-rl with importance weighted actor-learner architectures},
  author={Espeholt, Lasse and Soyer, Hubert and Munos, Remi and Simonyan, Karen and Mnih, Volodymir and Ward, Tom and Doron, Yotam and Firoiu, Vlad and Harley, Tim and Dunning, Iain and others},
  journal={arXiv preprint arXiv:1802.01561},
  year={2018}
}

@inproceedings{bousmalis2018using,
  title={Using simulation and domain adaptation to improve efficiency of deep robotic grasping},
  author={Bousmalis, Konstantinos and Irpan, Alex and Wohlhart, Paul and Bai, Yunfei and Kelcey, Matthew and Kalakrishnan, Mrinal and Downs, Laura and Ibarz, Julian and Pastor, Peter and Konolige, Kurt and others},
  booktitle={2018 IEEE international conference on robotics and automation (ICRA)},
  pages={4243--4250},
  year={2018},
  organization={IEEE}
}

@misc{baselines,
  author = {Dhariwal, Prafulla and Hesse, Christopher and Klimov, Oleg and Nichol, Alex and Plappert, Matthias and Radford, Alec and Schulman, John and Sidor, Szymon and Wu, Yuhuai and Zhokhov, Peter},
  title = {OpenAI Baselines},
  year = {2017},
  publisher = {GitHub},
  journal = {GitHub repository},
  howpublished = {\url{https://github.com/openai/baselines}},
}

\newpage

\section{Appendix}
\begin{appendix}
\section{Search Spaces}
\label{app:search-spaces}
Table \ref{search_spaces_table} describes two search spaces that we used in our experiments. An abbreviation \{$1, ..., n$\} means that a parameter can vary from 1 to $n$ in a search space.
\begin{table*}[!h]
\begin{multicols}{2}
    \begin{center}
    \textbf{Search Space 1}
    \begin{tabularx}{0.5\textwidth} {@{}lcccc@{}}
        \hline
        \textbf{Layer} & \textbf{input} & \textbf{output} & \textbf{kernel} & \textbf{stride}\\
        \hline
        Conv-1 & $4$ & $32$ & $8$ & $4$         \\
        Conv-2 & $32$ & $64$ & $1, .., 5$ & $2$ \\
        Conv-3 & $64$ & $64$ & $1, .., 5$ & $1$ \\
        Padding & $-$ & $121*64$ & $-$ & $-$    \\        
        Linear & $121*64$ & $512$ & $-$ & $-$   \\
        \hline

    \end{tabularx}
    \end{center}

    \vfill\eject
    
    \begin{center}
    \textbf{Search Space 2}
    \begin{tabularx}{0.5\textwidth}{@{}lcccc@{}}
        \hline
        \textbf{Layer} & \textbf{input} & \textbf{output} & \textbf{kernel} & \textbf{stride}\\
        \hline
        Conv-1 & $4$ & $32$ & $8$ & $4$         \\
        Conv-2 & $32$ & $64$ & $2, 5$ & $1$     \\
        Conv-3 & $64$ & $64$ & $2, 5$ & $1$     \\
        Conv-4 & $64$ & $64$ & $2, 5$ & $1$     \\
        Conv-5 & $64$ & $64$ & $2, 5$ & $1$     \\
        Conv-6 & $64$ & $64$ & $2, 5$ & $1$     \\
        Padding & $-$ & $121*64$ & $-$ & $-$    \\        
        Linear & $121*64$ & $512$ & $-$ & $-$   \\
        \hline
    
    \end{tabularx}
    \end{center}
\end{multicols}
\caption{Detailed description of the search spaces.}
\label{search_spaces_table}
\end{table*}

\section{Nature CNN architecture}
\label{app:nature-CNN-architecture}
Table \ref{tab:nature-cnn} describes the architecture ``Nature CNN'' \cite{Humanlevel2015} which is a member of the  \textbf{search space 1}.
\begin{table}[!h]
\begin{center}
\begin{tabularx}{0.5\textwidth}{@{}lcccc@{}}
    \hline
    \textbf{Layer} & \textbf{input} & \textbf{output} & \textbf{kernel} & \textbf{stride}\\
    \hline
        Conv-1 & $4$ & $32$ & $8$ & $4$         \\
        Conv-2 & $32$ & $64$ & $4$ & $2$        \\
        Conv-3 & $64$ & $64$ & $3$ & $1$        \\
        Padding & $-$ & $121*64$ & $-$ & $-$    \\
        Linear & $121*64$ & $512$ & $-$ & $-$   \\
    \hline
\end{tabularx}
\caption{Convolution network architecture that used in \cite{Humanlevel2015} that's contain 3 convolution layer followed by flatten and linear layers.}\label{tab:nature-cnn}
\end{center}
\end{table}

\section{Implementation details}
\label{app:hyperparams}
In this section, we present the hyperparameters used for training the RL agents in the ATARI games environment (Table \ref{tab:params_scratch}), as well as the hyperparameters used for ENAS and SPOS (Table \ref{tab:params_enas}).
We use the same set of parameters for both training from scratch experiments, and training the child networks sampled by the NAS controllers.

\begin{table}[!h]
\begin{center}
\begin{tabularx}{0.6\textwidth}{c|c}
    \hline
    \textbf{Hyperparameter's name} & \textbf{Value}\\
    \hline
        \# timesteps & $10$M    \\
        \# runner timesteps & $128$   \\  
        $\epsilon$ (PPO) & $0.1$   \\
        value loss coef. (PPO) & $0.25$ \\
        $\lambda$ (GAE) & $0.95$   \\
        entropy coef. & $0.01$   \\
        learning rate & CosineAnnealing($0.00025$)   \\
        \# parallel env. & $8$  \\
              
    \hline
\end{tabularx}
\caption{The hyperparameters for training PPO agents on ATARI games.}
\label{tab:params_scratch}
\end{center}
\end{table}

\begin{table}[!h]
\begin{center}
\begin{tabularx}{0.6\textwidth}{c|c}
    \hline
    \textbf{Hyperparameter's name} & \textbf{Value}\\
    \hline
        \# child network epochs & $10$    \\
        \# NAS runner timesteps & $3$   \\  
        NAS entropy coef. & $0.0001$   \\
        NAS learning rate & CosineAnnealing($0.001$)   \\
        baseline momentum. & $0.2$  \\
              
    \hline
\end{tabularx}
\caption{The hyperparameters for ENAS training.}
\label{tab:params_enas}
\end{center}
\end{table}

\section{The best architectures}
\label{app:found-architectures}
Tables \ref{tab:best-1}, \ref{tab:best-2}, \ref{tab:best-3}, \ref{tab:best-4} show the best architectures for each game.  The architectures tend to be similar for both of the search spaces except the kernel size for the last layers.
\begin{table}[!h]
\begin{center}
\begin{tabularx}{0.5\textwidth}{@{}lcccc@{}}
    \hline
    \textbf{Layer} & \textbf{input} & \textbf{output} & \textbf{kernel} & \textbf{stride}\\
    \hline
        Conv-1 & $4$ & $32$ & $8$ & $4$         \\
        Conv-2 & $32$ & $64$ & $4$ & $2$        \\
        Conv-3 & $64$ & $64$ & $5$ & $1$        \\
        Padding & $-$ & $121*64$ & $-$ & $-$    \\        
        Linear & $121*64$ & $512$ & $-$ & $-$   \\
    \hline
\end{tabularx}
\caption{The best architecture for Breakout extracted from search space 1 by SPOS using reward mean criteria.}\label{tab:best-1}
\end{center}
\end{table}

\begin{table}[!h]
\begin{center}
\begin{tabularx}{0.5\textwidth}{@{}lcccc@{}}
    \hline
    \textbf{Layer} & \textbf{input} & \textbf{output} & \textbf{kernel} & \textbf{stride}\\
    \hline
        Conv-1 & $4$ & $32$ & $8$ & $4$         \\
        Conv-2 & $32$ & $64$ & $3$ & $2$        \\
        Conv-3 & $64$ & $64$ & $2$ & $1$        \\
        Padding & $-$ & $121*64$ & $-$ & $-$    \\        
        Linear & $121*64$ & $512$ & $-$ & $-$   \\
    \hline
\end{tabularx}
\caption{The best architecture for Freeway extracted from search space 1 by SPOS using reward mean criteria.}\label{tab:best-2}
\end{center}
\end{table}

\begin{table}[!h]
\begin{center}
\begin{tabularx}{0.5\textwidth}{@{}lcccc@{}}
    \hline
    \textbf{Layer} & \textbf{input} & \textbf{output} & \textbf{kernel} & \textbf{stride}\\
    \hline
        Conv-1 & $4$ & $32$ & $8$ & $4$         \\
        Conv-2 & $32$ & $64$ & $5$ & $1$        \\
        Conv-3 & $64$ & $64$ & $5$ & $1$        \\
        Conv-4 & $64$ & $64$ & $5$ & $1$        \\
        Conv-5 & $64$ & $64$ & $2$ & $1$        \\
        Conv-6 & $64$ & $64$ & $2$ & $1$        \\
        Padding & $-$ & $121*64$ & $-$ & $-$    \\        
        Linear & $121*64$ & $512$ & $-$ & $-$   \\
    \hline
\end{tabularx}
\caption{The best architecture for Breakout extracted from search space 2 by SPOS using reward mean criteria.}\label{tab:best-3}
\end{center}
\end{table}

\begin{table}[!h]
\begin{center}
\begin{tabularx}{0.5\textwidth}{@{}lcccc@{}}
    \hline
    \textbf{Layer} & \textbf{input} & \textbf{output} & \textbf{kernel} & \textbf{stride}\\
    \hline
        Conv-1 & $4$ & $32$ & $8$ & $4$         \\
        Conv-2 & $32$ & $64$ & $5$ & $1$        \\
        Conv-3 & $64$ & $64$ & $5$ & $1$        \\
        Conv-4 & $64$ & $64$ & $5$ & $1$        \\
        Conv-5 & $64$ & $64$ & $5$ & $1$        \\
        Conv-6 & $64$ & $64$ & $5$ & $1$        \\
        Padding & $-$ & $121*64$ & $-$ & $-$    \\        
        Linear & $121*64$ & $512$ & $-$ & $-$   \\
    \hline
\end{tabularx}
\caption{The best architecture for Freeway extracted from search space 2 by SPOS using reward mean criteria.\\
}\label{tab:best-4}
\end{center}
\end{table}

\end{appendix}

\end{document}